\documentclass[10pt,twocolumn,letterpaper]{article}
\usepackage{subfigure}
\usepackage[pagenumbers]{iccv} %

\newcommand{\p}[1]{Aff-Grasp}
\newcommand{\m}[1]{GAT}
\newcommand{\mfull}[1]{Geometry-guided Affordance Transformer}
\newcommand{\ad}[1]{AED}

\newif\ifdrafting 
\draftingtrue %
\draftingfalse %

\draftingtrue

\ifdrafting
    \newcommand{\ls}[1]{{\color{pink}#1}}
\else
    \newcommand{\ls}[1]{}
\fi

\definecolor{iccvblue}{rgb}{0.21,0.49,0.74}
\usepackage[pagebackref,breaklinks,colorlinks,allcolors=iccvblue]{hyperref}
\usepackage{overpic}
\usepackage{makecell}
\usepackage{multirow}
\usepackage{subcaption}

\def\paperID{9971} %
\def\confName{ICCV}
\def\confYear{2025}

\title{Learning Precise Affordances from Egocentric Videos for Robotic Manipulation}

\author{Gen Li\textsuperscript{1} \quad Nikolaos Tsagkas\textsuperscript{1} \quad Jifei Song\textsuperscript{2} \quad Ruaridh Mon-Williams\textsuperscript{1} \quad Sethu Vijayakumar\textsuperscript{1} \\ Kun Shao\textsuperscript{2}\footnotemark[1] \quad Laura Sevilla-Lara\textsuperscript{1}\thanks{Corresponding authors} \\ \\ 
\textsuperscript{1}University of Edinburgh\quad
\textsuperscript{2}Huawei Noah's Ark Lab\quad
\vspace{-0.3cm}
}

\begin{document}
\maketitle

\begin{abstract}
Affordance, defined as the potential actions that an object offers, is crucial for embodied AI agents.
For example, such knowledge directs an agent to grasp a knife by the handle for cutting or by the blade for safe handover.
While existing approaches have made notable progress, affordance research still faces three key challenges: data scarcity, poor generalization, and real-world deployment.
Specifically, there is a lack of large-scale affordance datasets with precise segmentation maps, existing models struggle to generalize across different domains or novel object and affordance classes, and little work demonstrates deployability in real-world scenarios.
In this work, we address these issues by proposing a complete affordance learning system that
(1) takes in egocentric videos and outputs precise affordance annotations without human labeling, 
(2) leverages geometric information and vision foundation models to improve generalization, and 
(3) introduces a framework that facilitates affordance-oriented robotic manipulation such as tool grasping and robot-to-human tool handover.
Experimental results show that our model surpasses the state-of-the-art by 13.8\% in mIoU, and the framework achieves 77.1\% successful grasping among 179 trials,
including evaluations on seen, unseen classes, and cluttered scenes.
Project page: \url{https://reagan1311.github.io/affgrasp}.
\end{abstract}

\section{Introduction}
\label{sec:intro}
\begin{figure}[t]
    \centering
    \includegraphics[width=0.475\textwidth]{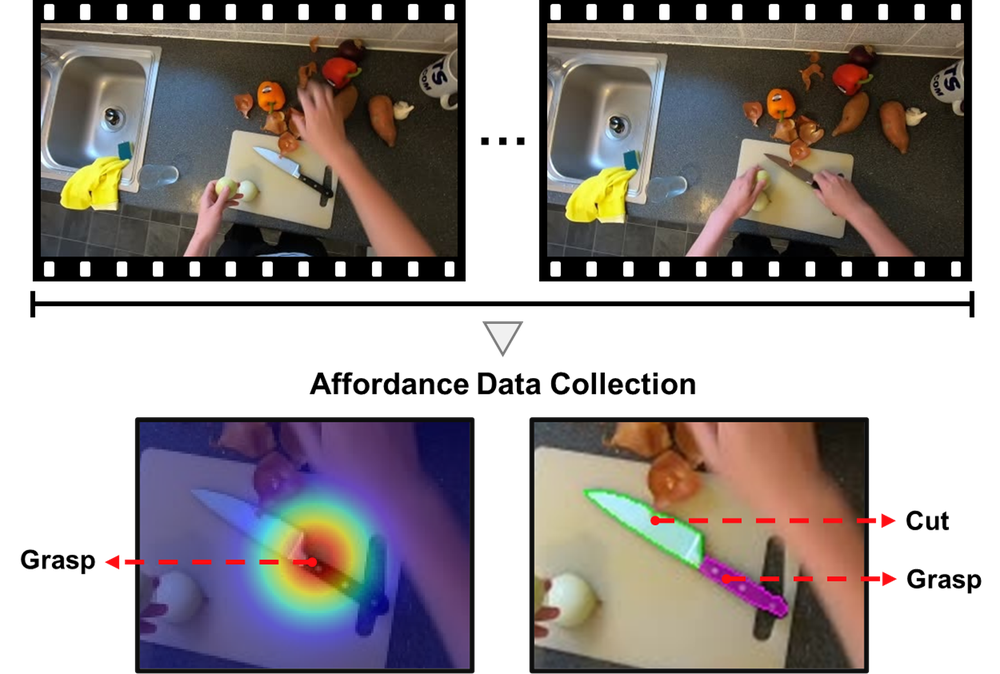}
    \caption{Illustration of affordance data collection and robot deployment. Existing work~\cite{forecast-hoi, joint_hand_hotspot, robo-abc, vrb} collects the graspable affordance as Gaussian heatmaps, whereas we extract both graspable and functional affordances with precise segmentation masks, enabling tool grasping, tool-object interaction, and robot-to-human tool handover.}\label{fig:teaser}
\end{figure}

Understanding affordances of an object means knowing the possible actions it enables as well as locating specific object parts for those actions. 
This knowledge serves as a key link between perception and action, enabling an intelligent system to transition from observing how an object is used to actually performing the action itself.
Despite its importance and recent progress, current affordance-related research often suffers from three major challenges: data scarcity, poor generalization, and real-world deployment.
First, affordance datasets are scarce and expensive to create, as they require detailed annotations of object parts, which are small, low-resolution, and often occluded.
For example, accurately segmenting the handle of a spoon for holding is challenging due to its thin structure and potential occlusions in cluttered scenes.
Second, current models often struggle to generalize to diverse new objects, affordances, or environments, as they are typically trained from scratch on a limited set of data. 
Finally, few affordance studies deploy methods in real robots, where the models need to be robust to noise, novel scenes, and other real-world factors.
These three challenges are in fact deeply interconnected: a lack of large-scale and diverse data limits model generalization, which is essential for reliable real-world deployment.

To tackle each of these challenges, we propose a comprehensive affordance learning system that encompasses data collection, model learning, and robot deployment.
The system starts with an automatic pipeline for collecting training data from egocentric videos, which are rich in human-object interactions.
While a number of studies~\cite{grounded, joint_hand_hotspot, vrb, robo-abc} have explored extracting affordances from egocentric videos, two limitations persist in their pipeline as illustrated in~\cref{fig:teaser}: 
(1) The focus is primarily on how humans grasp objects (graspable affordance), rather than on which part of the tool is being used (functional affordance).
(2) Affordances are learned and represented as probabilistic distributions, which are coarse and noisy, making them difficult to apply in real-life situations and susceptible to distractions.
To resolve these limitations, we aim to jointly annotate both graspable (\eg, object handles) and functional affordances (\eg, knife blades, hammerheads), focusing on generating precise segmentation maps rather than coarse heatmaps.
Concretely, given an egocentric video, our pipeline first extracts graspable points on objects from hand-object interactions and functional points from tool-object interactions. 
To cope with occlusions, we identify the pre-contact frame, where contact is about to occur, 
and project the extracted points to this frame through homography or point correspondence.
Finally, these points act as prompts for the segment anything model (SAM)~\cite{sam} to obtain precise part segmentation.

Although the previous step enables collection of images and annotations at a low cost, the resulting data pose challenges for training.
Most samples have quite low resolution, with cropped areas often comprising only 5\% of the original frame. 
To address this issue, we propose~\mfull{} (\m{}), which enriches and constrains the prediction process by leveraging the geometric features, rather than relying solely on blurry or low-resolution appearance.
Moreover, we observe that the model yields inferior performance when evaluated on data that largely differs from the training source domain.
To tackle this domain gap, we use the visual foundation model DINOv2~\cite{dinov2} as the image encoder, which has been trained on data from various domains.
Lastly, to enable affordance-oriented manipulation, we introduce \p{}, a framework that combines the \m{} with a grasp generation model for robotic manipulation. 
Building on the accurate predictions of graspable and functional affordances by \m{}, \p{} can handle tasks beyond simple grasping, including tool-object interaction and robot-to-human tool handover (see examples in~\cref{fig:teaser}).

To comprehensively demonstrate the effectiveness of our methods, we perform evaluations from two perspectives. 
First, we collect and annotate images from several existing affordance datasets and internet sources, creating a challenging evaluation dataset of great diversity to assess the model's performance. 
Second, we design a real-world robotic manipulation evaluation with 7 tasks and 34 diverse objects. 
It is worth nothing that both evaluations include out-of-domain data and novel objects, making this a challenging cross-domain and zero-shot setup.

Overall, the contributions of this work can be summarized as follows:
(1) \textit{Automated Affordance Data Collection:} We propose an automated pipeline for collecting and annotating affordance data from egocentric human-object interaction videos.
Different from previous work, the data are collected with precise segmentation maps for both graspable and functional affordances. 
(2) \textit{Advanced Affordance Learning Model:} 
We introduce \mfull{} (\m{}) that incorporates shape and geometric priors in a flexible and innovative way to tackle the challenging affordance segmentation task.
(3) \textit{Affordance-Oriented Manipulation:} We present \p{}, a framework designed for affordance-oriented grasping.
\p{} can identify the most suitable object based on task instructions, grasp the appropriate part, and utilize its functional component to complete the task (without specifying explicit object or part names).
(4) \textit{Extensive Vision and Robot Evaluations:} We conduct experiments on both visual data and a real robot.
A challenging affordance evaluation dataset is created for vision evaluation and diverse tasks are designed for robot experiments across a wide range of objects.

\section{Related Work}
\begin{figure*}[ht]
\centering
\includegraphics[width=.99\textwidth]{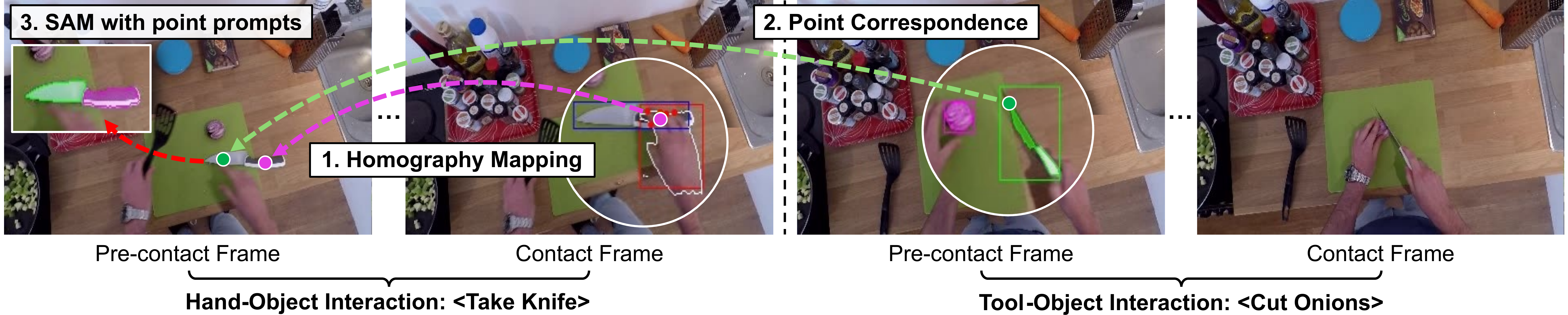}
\vspace*{-2mm}
\caption{Illustration of the data collection process from egocentric videos. First, graspable points (depicted in purple) are localized from clips of hand-object interaction and then projected to pre-contact frame by homography. Next, functional points (depicted in green) are identified from tool-object interactions and mapped to the pre-contact frame of hand-object interaction through point correspondence. Lastly, these points are used as prompts for the SAM to obtain affordance masks.}\label{point-fig}
\end{figure*}

\noindent \textbf{Learning Affordance from Human Videos.}~An emerging and promising alternative for fully supervised affordance learning is the automatic extraction of affordance knowledge through the observation of human interactions.
Given the abundance of human-object interaction video datasets~\cite{epic,epic_v2,Damen2018EPICKITCHENS,oh2019video,ego4d,Li_2018_ECCV,hoi4d,oakink,oakink2}, many studies~\cite{demo2vec, forecast-hoi, joint_hand_hotspot, vrb, robo-abc} have explored how to extract rich affordance information from videos.
Despite making significant strides in affordance learning, these studies have two limitations.
First, the defined affordance is limited to graspable areas of objects, ignoring the important functional parts.
Second, the affordance is predicted as coarse heatmaps, which are less accurate when applied in the real world.
While some recent work \cite{multi-label-aff, wacv-aff, text-aff} has made attempts to address these limitations, none have tackled both aspects simultaneously.
To bridge these gaps, we introduce an improved affordance data collection pipeline that localizes both graspable and functional points, and utilizes SAM~\cite{sam} to produce high-quality masks with point-based prompting.

\vspace{1mm}
\noindent\textbf{Affordance-Oriented Robotic Manipulation.}~Recent work~\cite{voxposer, lerf-togo, wang2023d3fields, shen2023F3RM,tsagkas2024click, copa, ellmer} has achieved task-oriented grasping using vision-language models (VLMs) and large language models (LLMs)~\cite{clip, dinov2, dino, stable-diff, gpt4v, gpt4}.
These methods use LLMs to infer parts to be grasped based on task instructions, and VLMs to localize the relevant regions.
While this pipeline can produce effective and accurate manipulation, the reasoning process often requires prompt engineering and can be time-consuming.
In contrast, there has been less focus on more efficient affordance-oriented grasping that can derive graspable and functional areas without explicitly specifying corresponding object parts.
One major barrier to this direction is the lack of large-scale affordance datasets, compounded by the difficulty of unifying existing ones due to inconsistent annotation standards.

To overcome the data limitations, recent methods such as VRB~\cite{vrb} and Robo-ABC~\cite{robo-abc} have explored the potential of learning affordances from human videos for robotic manipulation.
However, VRB generates only coarse Gaussian heatmaps, and Robo-ABC relies on point correspondences that can be noisy and susceptible to background variations.
In contrast, we propose a more effective and robust strategy to extract precise affordance masks, and introduce an efficient model for affordance learning.
When combined with a small set of basic pre-recorded motion primitives (as in~\cite{dipalo2024on, dipalo2024dinobot}) and grasp pose detection models~\cite{contact-grasp, anygrasp, graspnet}, our method enables the utilization of both graspable and functional affordances.

\section{Method}
In this work, our goal is to develop a holistic system that covers data collection, model learning, and robot deployment. 
To this end, we first develop an automated pipeline to collect images and related affordance annotations from human videos.
Next, we propose an effective affordance learning model termed \mfull{} (\m{}).
Finally, we introduce \p{} that couples the trained model with an off-the-shelf grasp estimation model to achieve affordance-oriented manipulation.
In~\cref{video}, we describe how affordance data are collected from large-scale egocentric videos of human interactions. 
We then elaborate on the design of \m{} that enables effective affordance learning from collected data in~\cref{model}.
Lastly, in~\cref{robot}, we explain \p{}, detailing how it yields affordance-oriented grasp poses.

\begin{figure*}[t]
\centering
\includegraphics[width=.95\textwidth]{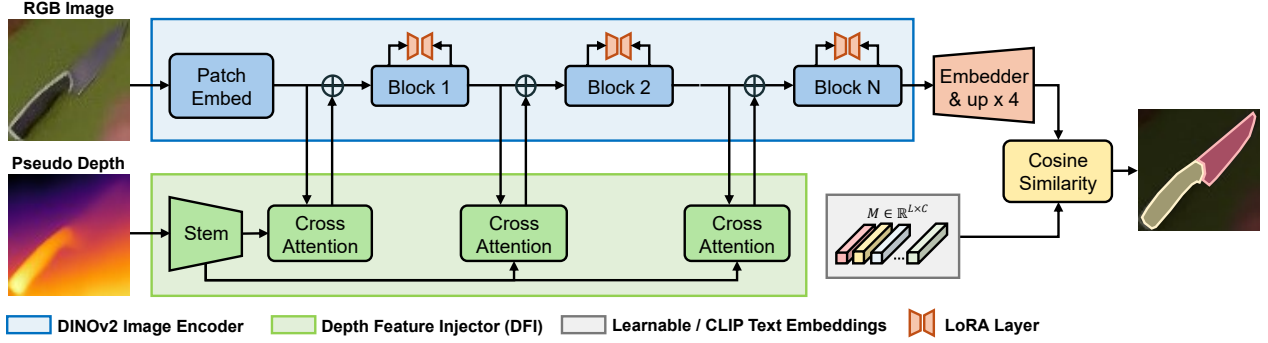}
\caption{The architecture of \m{}. It consists of a DINOv2 image encoder, a depth feature injector, an embedder, and LoRA layers. The model performs segmentation by computing cosine similarity between upsampled features and learnable / CLIP text embeddings.
}\label{model-fig}
\end{figure*}

\subsection{Data Collection from Egocentric Videos}\label{video}
Given an egocentric video of a human interacting with an object, our aim is to first locate contact points.
Human-object interaction videos can generally be categorized into two types: hand-object interaction and tool-object interaction.
In hand-object interaction, contact points indicate where the human grasps the object.
In tool-object interaction, contact points reveal which part of the tool is used to interact with the target object.
These points represent sparse \textit{graspable} and \textit{functional} areas of an object, carrying rich affordance information.
As shown in~\cref{point-fig}, we propose a pipeline to automatically collect these points without manual annotation.
The collected points then serve as prompts to produce precise segmentation masks using SAM~\cite{sam}.

\vspace{1mm}
\noindent\textbf{Graspable Point Localization.}~Egocentric videos like Epic-Kitchens~\cite{epic} and Ego4D~\cite{ego4d} include timestamped narrations that describe actions and their respective start and end times.
Based on these narrations, we first retrieve hand-object interaction clips (associated with actions such as ``take'' or ``hold'') and employ a hand-object detector~\cite{ho-detector} to generate contact states and hand-object bounding boxes for all frames.
Next, we use labeled timestamps to extract the contact frame, which is typically annotated as the start of an action.
We conduct an additional hand segmentation in this frame using the hand box as a prompt via EfficientSAM~\cite{eff-sam}.
We then locate the intersection region of the hand mask and object bounding box to sample $n$ contact points $P=\{p_{1}, p_{2}, ..., p_{n}\}$.
However, the sampled points are often occluded by hands, and therefore do not accurately represent the graspable affordance area of the object.
To collect clean object images free of occlusion, it is necessary to identify the pre-contact frame, \ie, the last frame where the object is fully visible before contact occurs.
We utilize the contact states to detect the frame that is closest to the contact frame but without hand-object contact, designating this as the pre-contact frame.
Since human motion between adjacent frames is minimal, we follow a similar pipeline to previous studies~\cite{joint_hand_hotspot, robo-abc} to project the average position of sampled graspable points to the pre-contact frame by computing a homography transformation~\cite{homography}, as illustrated by the purple dashed line in~\cref{point-fig}.

\vspace{1mm}
\noindent\textbf{Functional Point Localization.}~To localize functional points, we first retrieve the relationship between objects and affordances from existing affordance datasets.
We then extract related tool-object interaction clips from video narrations based on the object-affordance relationship.
For example, most datasets associate affordances ``cut'' and ``grasp'' with a knife.
After localizing the graspable points from a clip showing a person grasping a knife, we then retrieve the following nearest clip depicting a cutting action with the knife.

However, the tool is often heavily occluded or invisible in the contact frame.
Similar to graspable point localization, we need to find the pre-contact frame that shows minimal or no intersection between the tool and the target object.
To achieve this, we first employ the same combination of the hand-object detector and EfficientSAM to obtain the bounding box and mask of the hand-held tool.
Next, we use an open-vocabulary object segmentation model, GroundedSAM~\cite{groundedsam}, to segment the target object.
We then measure the Intersection over Union (IoU) between tool and object bounding boxes in frames prior to the contact frame until the IoU is below a preset threshold.
Lastly, we calculate the point distances between masks in this pre-contact frame and extract the point within the tool mask that has the shortest distance to all points in the object mask.
An erosion operation is applied to the tool mask to ensure that the functional point is inside the tool.

Nonetheless, not all object categories have related action clips in the narrations.
In such cases, we use the farthest sampling to determine the functional points based on the distance to the grasp points.
This simple method produces accurate functional points, as most tools are designed with graspable and functional parts distributed at opposite ends.

\vspace{1mm}
\noindent \textbf{Data Generation.}~After extracting functional points, we first project these points to the pre-contact frame of the hand-object interaction clip where we infer the graspable points.
Since the object category remains the same, we compute the point correspondence within object bounding boxes using foundation model features~\cite{deepvit}, which map the functional point from the tool-object pre-contact frame to the hand-object pre-contact frame (illustrated by the green dashed line in~\cref{point-fig}).
We then label the graspable points as positive and the functional points as negative to obtain the graspable affordance mask.
Conversely, the functional affordance mask is generated by reversing the positive and negative labels.
Finally, we crop the object images and store them along with the generated segmentation masks as annotations.

\subsection{\mfull{}}\label{model}

While the affordance data are collected from egocentric videos without manual labor, there are two issues that hinder effective model training.
The first issue is the low resolution of the collected images. The object of interest occupies very small areas of the video frames, often resulting in cropped images of less than 100 pixels in either height or width.
The second is the limited diversity of the training data, characterized by monotonous backgrounds and mostly restricted to indoor scenes.
To cope with these issues, we propose an affordance learning architecture called \m{}, as illustrated in~\cref{model-fig}.
It includes a novel Depth Feature Injector (DFI) that integrates geometric priors into image features using pseudo depth maps, a DINOv2 image encoder as a feature extractor, and additional LoRA layers for effective fine-tuning.

\vspace{1mm}
\noindent \textbf{Depth Feature Injector.}~We argue that depth maps introduce rich geometric information that can help with foreground-background and part separation.
Additionally, they exclude color information, allowing the model to fully focus on the shape information, which is highly relevant to affordances.
For instance, graspable parts often consist of shapes like cylinders or spheres, while parts designed for cutting typically feature sharp edges and flat surfaces.

Specifically, we first obtain pseudo depth maps for each training image with a state-of-the-art depth estimation model Depth-Anything~\cite{depth-any}.
During model training, the pseudo depth map is first encoded into feature maps using a stem block, which contains three standard $3\times3$ convolution layers.
These feature maps are then processed by a $1\times1$ convolution that transforms the channel dimension to match that of the RGB image features.

We divide the whole model into four blocks.
At the beginning of each block, the DFI takes the image features $F_i\in\mathbb{R}^{N\times C}$ and depth features $F_{d} \in\mathbb{R}^{N\times C}$ as input, and outputs updated image features $\hat{F}_i\in\mathbb{R}^{N\times C}$, where $N$ denotes the number of patches.
Concretely, DFI contains several cross-attention layers followed by residual connections.
In the cross-attention layer, $F_i$ is used as the query, and $F_d$ is adopted as the key and value:
\begin{gather}
Q=\phi_{q}(F_i), ~K=\phi_{k}(F_d), ~V=\phi_{v}(F_d),
\\
\hat{F}_i = \beta\cdot\text{softmax}(QK^T/\sqrt{d_k})\cdot V + F_i,
\end{gather}
where $\phi$ is a linear transformation, and $d_k$ is the dimension of the key acting as a scaling factor. Following \cite{vit-adapter}, we set a learnable vector $\beta\in\mathbb{R}^{C}$, initialized to 0, to balance the output from the cross-attention layer and the image feature.
This strategy prevents the image feature from being excessively affected by the depth feature, making the training process more stable.
We observe that DFI constantly brings improvement, even when integrated solely during training (see~\cref{sec:ablation}).
This indicates that it can act as a regularization mechanism during training, and can be discarded during inference to speed up the process.

\vspace{1mm}
\noindent \textbf{DINOv2 with Low-Rank Adaptation.}~We notice that directly training from the typical ImageNet pre-trained representation often leads to inferior results.
This can be attributed to two primary reasons: First, affordance segmentation focuses on fine-grained object parts, whereas representations trained for image classification emphasize global object features~\cite{deepvit}.
Second, ImageNet pre-trained models exhibit limited diversity, making it challenging to handle data from diverse domains.
To address this issue, we employ the self-supervised visual foundation model DINOv2, which has been demonstrated to be highly effective for data-limited affordance learning~\cite{ooal}.
Furthermore, we introduce LoRA~\cite{lora} to fine-tune the model without modifying the parameters of the original DINOv2.
This strategy helps adaptation across different domains and prevents overfitting.
LoRA was originally developed to fine-tune large language models for different downstream tasks.
Specifically, it injects trainable rank decomposition matrices to a pre-trained weight matrix $W_0\in\mathbb{R}^{d\times k}$ by $W_0 + \Delta W = W_0 + BA$, where $B\in\mathbb{R}^{d\times r}$, $A\in\mathbb{R}^{r\times k}$, and the rank $r \ll \min(d, k)$.
During training, only $A$ and $B$ are trainable, while $W_0$ remains frozen. This incurs minimal computational cost and memory usage.

\begin{figure*}[t]
\centering
\includegraphics[width=0.85\textwidth]{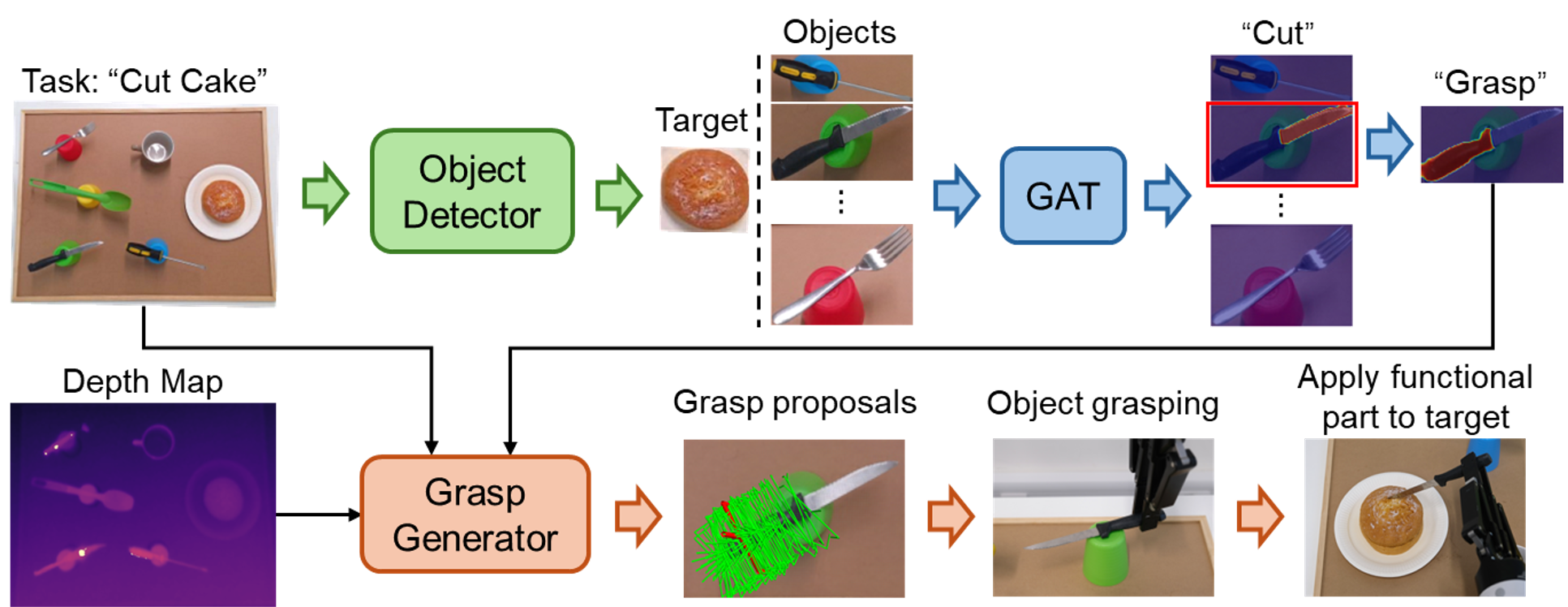}
\caption{The framework of \p{}. It first employs an open-vocabulary detector~\cite{groundingdino} to locate all objects within the scene, which are then sent to \m{} to determine if they possess the corresponding affordance required for the task. Afterwards, a 6 DoF grasp generation model, Contact-GraspNet, leverages the object's graspable affordance and the depth map to generate dense grasp proposals. Finally, the robot executes affordance-specific sequential motion primitives to apply the functional part to the target.
}\label{grasp-pipeline}
\end{figure*}

\vspace{1mm}
\noindent \textbf{Classifier and Loss Functions.}~To speed up inference time for real-world applications, we avoid adding any complex decoder structures.
Instead, we process the output feature with an embedder (an MLP), reshape it, and upsample it by a factor of four to increase the resolution $F_{out}\in\mathbb{R}^{C\times \frac{4H}{p}\times \frac{4W}{p}}$, where $p$ is the patch size.
Next, we intialize $M\in\mathbb{R}^{{L\times C}}$ learnable embeddings, where $L$ is the number of affordance categories.
We compute the cosine similarity between $M$ and $F_{out}$ to yield the segmentation output, which is then restored to the same size as the input image via bilinear interpolation.
Due to the domain gap, we do not add a learnable embedding for the background classification to prevent overfitting.
Alternatively, we determine a pixel as background if all its affordance predictions are below a preset threshold $\tau$.
Compared to a linear layer and explicit background classifier, the cosine similarity-based segmentation and implicit background prediction are more robust and can effectively improve the performance, as detailed in~\cref{sec:ablation}.
In addition, to achieve open vocabulary affordance segmentation, $M$ can also be replaced with corresponding CLIP text embeddings, as verified in \cite{ooal}.

Since the collected data are highly unbalanced, we utilize a combination of focal loss~\cite{focal} and dice loss~\cite{dice} as training objectives:
\begin{gather}
\mathcal{L} = \alpha\cdot \mathcal{L}_{focal} + \mathcal{L}_{dice},
\end{gather}
where $\alpha$ is a weighting factor to balance loss values.

\subsection{Affordance-Oriented Robotic Manipulation}\label{robot}
Our ultimate goal is affordance-oriented robotic manipulation, where given a task and a cluttered scene, the robot can select the object that possesses the related affordance, grasp the correct part, and apply the functional part to the target object to perform desired actions.
To achieve this, we propose \p{}, which integrates \m{} to achieve affordance segmentation and transforms the visual affordance to available grasp poses.
The framework of \p{} is shown in~\cref{grasp-pipeline}.
Given a task consisting of a verb and a target, such as ``cut cake'', it first uses an open-vocabulary object detection model~\cite{groundingdino} to detect the target (cake) and other visible objects.
For objects other than the target, the input vocabulary is simply set to ``objects'' for class-agnostic detection.
These detected objects are then cropped and sent to \m{} to predict affordance areas.
The object with the most certain affordance area for the required action (cut) is identified.
After that, we extract the graspable affordance area of this object, \ie, the knife handle, and generate potential grasp poses within it.
For grasp pose estimation, we select Contact-GraspNet~\cite{contact-grasp} that can produce dense grasp proposals within the specified mask area.
The one with the highest score is then chosen as the execution pose.
Once the object is grasped and lifted, we execute affordance-specific sequential motion primitives to apply the functional affordance area to the target object to complete the task.
For the handover task, \p{} is instructed to find available grasp proposals within the functional affordance area and then pass the graspable part to the human hand.
When CLIP text embeddings are used as classifiers, the action required for a task can be transformed into text embeddings for open-vocabulary affordance segmentation~\cite{ov-aff-3d, ooal}.
Therefore, unseen affordance vocabularies can also be used at inference, enhancing the model's adaptability and versatility.

\section{Experiments}
In this section, we present experiments from both vision and robot perspectives, along with ablation studies that examine the design choices of \m{}.
Note that all evaluations are conducted in a zero-shot or cross-domain setting, since the test data and robot setup contain novel objects or affordances, and differ greatly from the training data collected from egocentric videos.
As a result, only models with strong generalization ability can achieve high performance.
Implementation details are provided in the supplementary.

\subsection{Vision Experiments}
\label{sec:vision_exp}

\textbf{Evaluation Dataset.}~To evaluate the effectiveness of \m{}, a diverse and challenging affordance dataset that has consistent object and affordance categories with the collected training data is needed.
After carefully inspecting existing datasets, we found that most of them are not compatible with our evaluation requirements.
Many datasets either have a small number of categories~\cite{aff_with_CNN_iit, learn2act_properly} or are collected in the lab environment with limited diversity~\cite{tool_parts_umd, cad120}.
Some datasets contain a large number of images but have coarse keypoint-based annotations~\cite{ag_from_exocentric_imgs, demo2vec} or small resolutions~\cite{aff_low_resolution_dataset}. 
Therefore, we create the Affordance Evaluation Dataset (\ad{}) by manually annotating 721 images collected from several existing affordance datasets and internet resources.
It contains 13 object categories and 8 affordance classes.
See supplementary for more details on AED.

\begin{table}[!t]
\centering
\resizebox{0.45\textwidth}{!}{
\begin{tabular}{clccc}
\toprule
\multicolumn{1}{c}{\textbf{Pre-train}}                                                                           & \textbf{Method}      & \textbf{mIoU} & \textbf{F1} & \textbf{Acc} \\ \midrule
\multicolumn{1}{c}{\multirow{3}{*}{\makecell{ImageNet}}}  & 

DeepLabV3+~\cite{deeplabv3+}  & \multicolumn{1}{c}{13.46}         & \multicolumn{1}{c}{22.27}         & 23.05    \\
&PSPNet~\cite{pspnet}      & \multicolumn{1}{c}{16.90}         & \multicolumn{1}{c}{27.32}         & 26.46     \\ 
& SegFormer~\cite{segformer} & \multicolumn{1}{c}{23.72}         & \multicolumn{1}{c}{36.86}         & 37.19     \\ 
\midrule

\multirow{5}{*}{\makecell{Foundation\\Models}}     
& ZegCLIP~\cite{zegclip}  & 18.33    &   26.41                           &    25.55   \\
& DINOv2~\cite{dinov2}    & 46.16                             & 62.49                             & 63.61                         \\
& ViT-Adapter~\cite{vit-adapter}    & 50.86                             & 66.88                             & 65.21                         \\
& OOAL~\cite{ooal}     &  54.82                            &   70.58                           &     68.00 \\
& \m{} (Ours)        & \textbf{68.62}           & \textbf{81.09}                             &       \textbf{83.51}         \\ \bottomrule        
\end{tabular}
}
\caption{Quantitative comparison on the \ad{}.}\label{tab:vision_com}
\vspace{-4mm}
\end{table}

\vspace{1mm}
\noindent \textbf{Quantitative and Qualitative Comparisons.}\label{sec:vis_com}~\Cref{tab:vision_com} shows the results of different segmentation approaches on the proposed \ad{}. 
They use either pre-trained ImageNet backbones to extract feature maps, or representations from visual foundation models like CLIP~\cite{clip} and DINOv2~\cite{dinov2}.
Thus, we categorize these models into two groups based on their pre-training strategies.
For ImageNet pre-trained models, we employ classical CNN segmentation models such as DeepLabV3+ and PSPNet, and a transformer-based model SegFormer.
For visual foundation-based models, we choose ZegCLIP, DINOv2, ViT-Adapter, and OOAL to compare with \m{}, as they represent the state-of-the-art in leveraging visual foundation models for semantic or affordance segmentation tasks. 
We notice that methods using pre-trained ImageNet backbones generally produce much inferior results compared to those based on foundation models.
This confirms the substantial domain gap between training and evaluation sources, and demonstrates that foundation models are more resistant to noisy training samples and have better cross-domain capabilities.

\begin{figure}[t]
\centering
\includegraphics[width=0.465\textwidth]{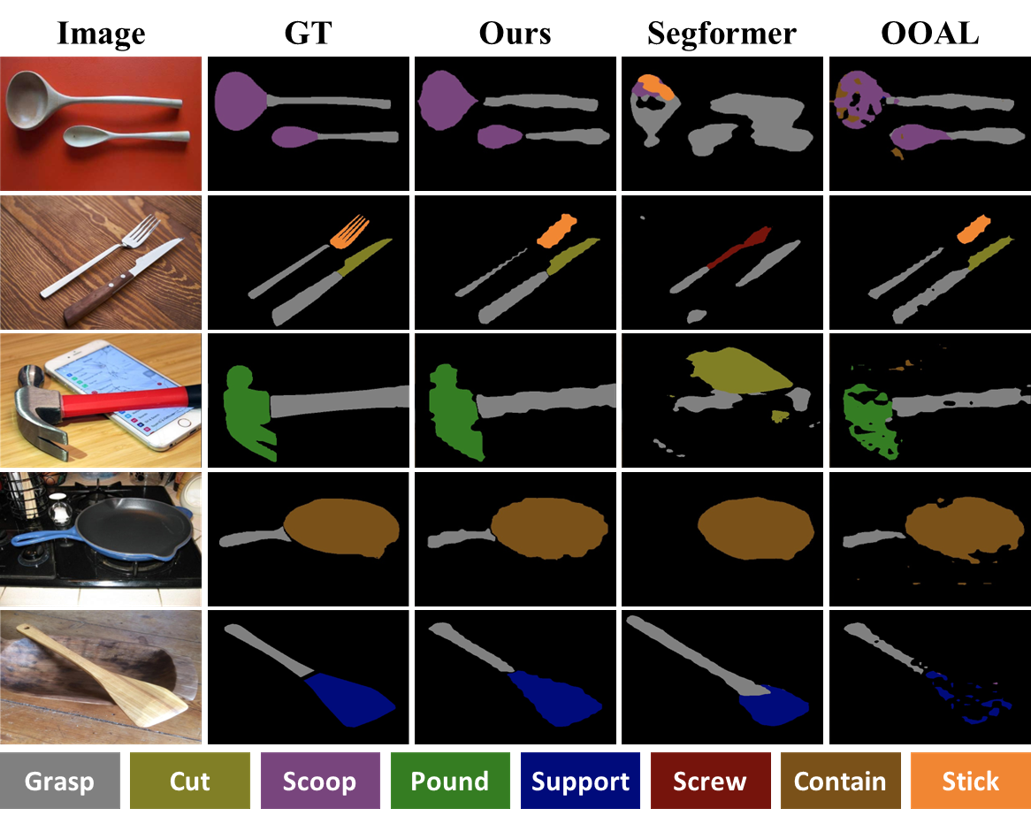}
\caption{Qualitative comparison between our approach and other segmentation models on the \ad{}.}\label{vis_com}
\end{figure}

\Cref{vis_com} depicts the qualitative comparison between our methods and other models. 
We find that models like PSPNet and Segformer often yield incomplete or incorrect affordance predictions, which may result from the low diversity in the pre-trained ImageNet representation.
On the other hand, OOAL can coarsely generate correct affordance prediction map, but often suffer from incomplete part activation and noisy segmentation around object boundaries.
In contrast, the results from \m{} are part-focused, and exhibit well-preserved boundaries.
Notably, our results stand out from other counterparts when dealing with images containing multiple objects.
Additional qualitative comparisons, as well as generalization tests on unseen objects and affordances are provided in the supplementary.

\subsection{Robot Experiments}
\label{sec:robot_exp}

\begin{figure}[t]
\centering
\includegraphics[width=0.45\textwidth]{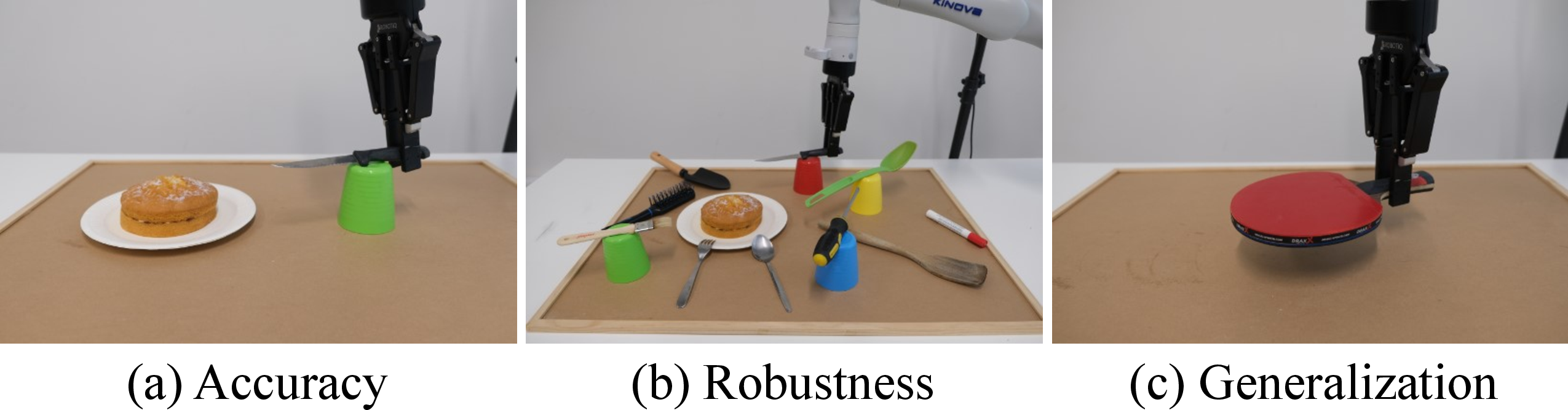}
\vspace{-2mm}
\caption{Illustration of accuracy, robustness, and generalization evaluations. (a) The accuracy evaluation requires the model to recognize the affordance of a single object and execute related tasks. (b) The robustness evaluation involves accurately selecting an object in a cluttered scene to perform a specified affordance task. (c) The generalization evaluation succeeds if the model can reason about the graspable area of unseen objects.}\label{eval-ill}
\end{figure}

\noindent \textbf{Experiment Setup and Comparison Methods.}~Real-world experiments are conducted to evaluate three essential properties: accuracy, robustness, and generalization. 
Illustrations for these experiments are shown in~\cref{eval-ill}.
For the comparison methods, since this work focuses on visual affordance learning (i.e., identifying appropriate graspable and functional regions) rather than grasp success, manipulation models are not directly comparable. Instead, we select two relevant methods, LOCATE~\cite{locate} and Robo-ABC~\cite{robo-abc}, that learn affordances in a similar manner---from egocentric images and videos, respectively.

\begin{table}[t]
\centering
\footnotesize
\resizebox{0.48\textwidth}{!}{
\begin{tabular}{lccc}
\toprule        Models & \makecell{Correct\\Affordance} & \makecell{Successful\\Grasp} & \makecell{Successful\\Interaction} \\ \midrule
LOCATE~\cite{locate}   &          42/72 (58.3\%)         &   33/72 (45.8\%)                &  n/a                     \\
Robo-ABC~\cite{robo-abc}   &          62/72 (86.1\%)         &   44/72 (61.1\%)                &  n/a                      \\
\p{} (Ours)  &          70/72 (97.2\%)         &   57/72 (80.6\%)                &  47/72 (65.3\%)                      \\ \bottomrule 
\end{tabular}
}
\caption{Success rates for accuracy evaluation. ``n/a" denotes inapplicability to tool-object interactions, as LOCATE's predictions often overlap, and Robo-ABC cannot infer functional areas.}\label{acc-tab}
\end{table}

\begin{table}[t]
\centering
\footnotesize
\resizebox{0.45\textwidth}{!}{
\begin{tabular}{lccc}
\toprule
Models         & \makecell{Correct\\Affordance} & \makecell{Successful\\Grasp} &  \makecell{Inference\\Time (s)} \\ \midrule
LOCATE~\cite{locate}   &    20/35 (57.1\%)               &   15/35 (42.9\%)    &    0.0047                       \\
Robo-ABC~\cite{robo-abc} &        24/35 (68.6\%)            &     21/35 (60.0\%)       & 12.92                       \\
\p{} (Ours)     &     32/35 (91.4\%)             &  28/35 (80.0\%)         &  0.0063           \\\bottomrule        
\end{tabular}
}
\caption{Success rates for generalization evaluation and inference time for affordance prediction components.}\label{gen-tab}
\end{table}

\vspace{1mm}
\noindent\textbf{Quantitative and Qualitative Comparisons.}
The results for the accuracy evaluation are shown in~\cref{acc-tab}.
It is clear that \p{} outperforms its competitors significantly, achieving an 11.1\% higher affordance prediction rate and a 19.5\% increase in successful grasping compared to Robo-ABC.
Also, it is worth mentioning that our success rate for affordance prediction is measured on both graspable and functional affordances, whereas other two methods are measured solely on the graspable affordance.
We observe that LOCATE struggles to make accurate predictions in real-world scenarios. 
Robo-ABC has relatively accurate affordance predictions, but it generates grasp proposals based on point correspondences, which do not represent the most confident grasp.
Consequently, even though Robo-ABC frequently makes correct affordance predictions, 29\% of its proposed grasping points do not lead to successful grasps.
In addition, we note that \p{} is capable of recognizing the correct affordance in cluttered scenes (robustness evaluation).
It achieves a high success rate in affordance prediction, accurately predicting affordances 95\% of the time, even in the presence of multiple seen and unseen objects acting as distractors (results and visual examples are provided in the supplementary).
Table~\ref{gen-tab} reports results from the generalization evaluation and the inference time for the affordance prediction component of the models.
It can be observed that our method is efficient and significantly more accurate in predicting the correct graspable areas for unseen objects, leading to a much higher success rate in grasping.

\begin{figure}[t]
\centering
\includegraphics[width=.45\textwidth]{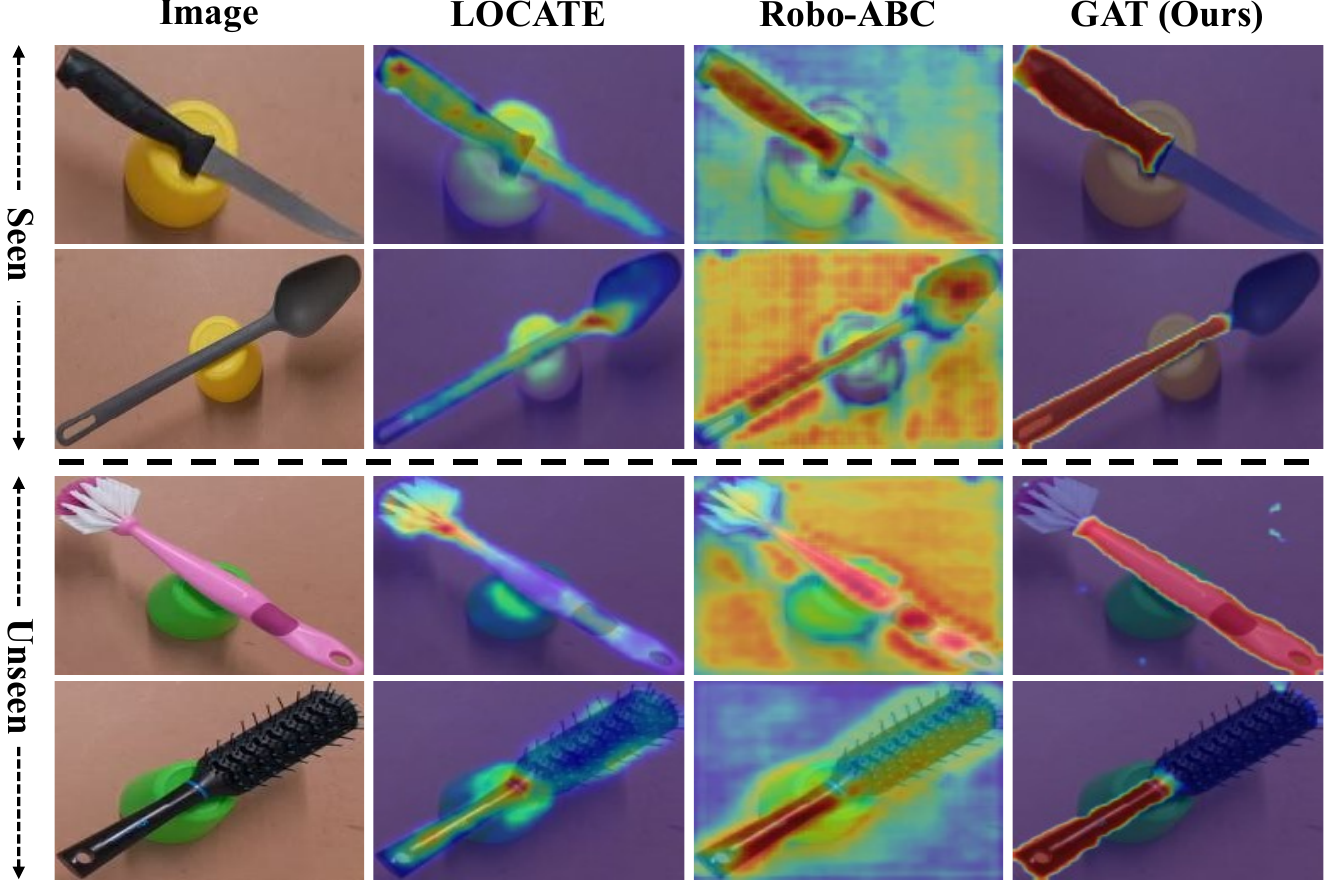}
\caption{Qualitative comparison of graspable affordance predictions on seen and unseen object categories.}\label{quali-com}
\end{figure}

In~\cref{quali-com}, we present raw predictions of graspable affordance from each model for seen and unseen objects.
It can be observed that LOCATE often produces incomplete and wrong predictions.
Robo-ABC occasionally makes predictions within the right object part area, but also produces high activation for the background or the entire object.
In comparison, \p{} consistently makes precise segmentation predictions for both seen and unseen objects and is not affected by the background.

\subsection{Ablation Study}
\label{sec:ablation}

\begin{table}[t]
\centering
\footnotesize
\begin{tabular}{lccc}
\toprule
             Methods                            & mIoU & F1 & Accuracy \\ \midrule
Baseline - DeiT III                               &  31.02    & 44.55   & 35.85     \\
w/ DINOv2                               &  45.45    & 61.78   & 70.86     \\
w/ embedder                               &  48.83    & 65.10   & 71.07    \\
w/ embedder \& up$\times$4 &  51.41    & 64.26   & 67.27    \\
w/ focal loss                             & 50.70     & 66.97   & 70.12    \\
w/ focal \& dice loss                     &  53.12    &  69.13  & 74.55    \\ \midrule

linear layer w/o bg  &  54.96    & 70.50   & 71.97    \\
cosine similarity                     &  55.52    & 71.01    &  71.54   \\
cosine similarity w/o bg  &  56.70    & 72.00   & 71.22    \\ \midrule

+ DFI-training only                & 60.15      & 74.92   & 79.87  \\
+ DFI                                     & 64.66    & 78.35   & 79.74    \\
+ LoRA                             & 68.62     & 81.09   & 83.51    \\ \bottomrule
\end{tabular}
\caption{Ablation results of embedder, loss functions, classifiers, and proposed modules. The baseline model is a DeiT III model with a linear layer and binary cross entropy loss. ``w/o bg'' means that there is no background classifier. ``DFI-training only'' denotes that the DFI is used during training, and discarded at inference.}\label{tab:ablation}
\end{table}

\begin{table}[t]
\centering
\resizebox{0.4\textwidth}{!}{
\begin{tabular}{lccc}
\toprule
Inference & \#Params (M) & GFLOPs & Time (ms) \\ \midrule
w/ DFI &  \phantom{000}96.9\textsubscript{\phantom{\textcolor{blue}{$\downarrow$5.4\%}}}    & \phantom{000}204.9\textsubscript{\phantom{\textcolor{blue}{$\downarrow$9.5\%}}}   & \phantom{000}10.1\textsubscript{\phantom{\textcolor{blue}{$\downarrow$37.6\%}}}   \\
w/o DFI &  \phantom{000}91.7\textsubscript{\textcolor{blue}{$\downarrow$5.4\%}}   & \phantom{000}185.5\textsubscript{\textcolor{blue}{$\downarrow$9.5\%}}   & \phantom{000}6.3\textsubscript{\textcolor{blue}{$\downarrow$37.6\%}}   \\
\bottomrule
\end{tabular}
}
\caption{Ablation study on inference efficiency of DFI.}\label{tab:ablation-dfi-efficiency}
\vspace{-2mm}
\end{table}

To explore the impact of each component in our model, we perform ablation experiments on the embedder, loss function, classifiers, and designed modules.
The ablation results are summarized in~\cref{tab:ablation}.
We first set up a baseline model, which employs a frozen DeiT III~\cite{deit3} backbone that is fully supervised on ImageNet.
We then add a simple linear layer for patch-wise classification and utilize binary cross-entropy as the loss function.
Based on this baseline, we first explore the impact of the embedder and loss functions.
We find that a larger feature map followed by an embedder is beneficial, and the combination of focal loss and dice loss also brings improvements.
Then, we analyze the results under different classification schemes, including the linear layer, cosine similarity, and whether to learn a background classifier.
It is clear that implicit background prediction leads to better performance.
Given the large gap between training and evaluation data, learning a background classifier can easily result in overfitting.
Also, employing cosine similarity as the classifier can better utilize the inherent features of DINOv2, producing better results than a linear classifier.
Lastly, we investigate the influence of DFI and LoRA.
Notably, DFI improves performance significantly by a large margin, with 7.96\% and 6.35\% increases in mIoU and F1 scores.
\begin{figure}[t]
\centering
\includegraphics[width=0.45\textwidth]{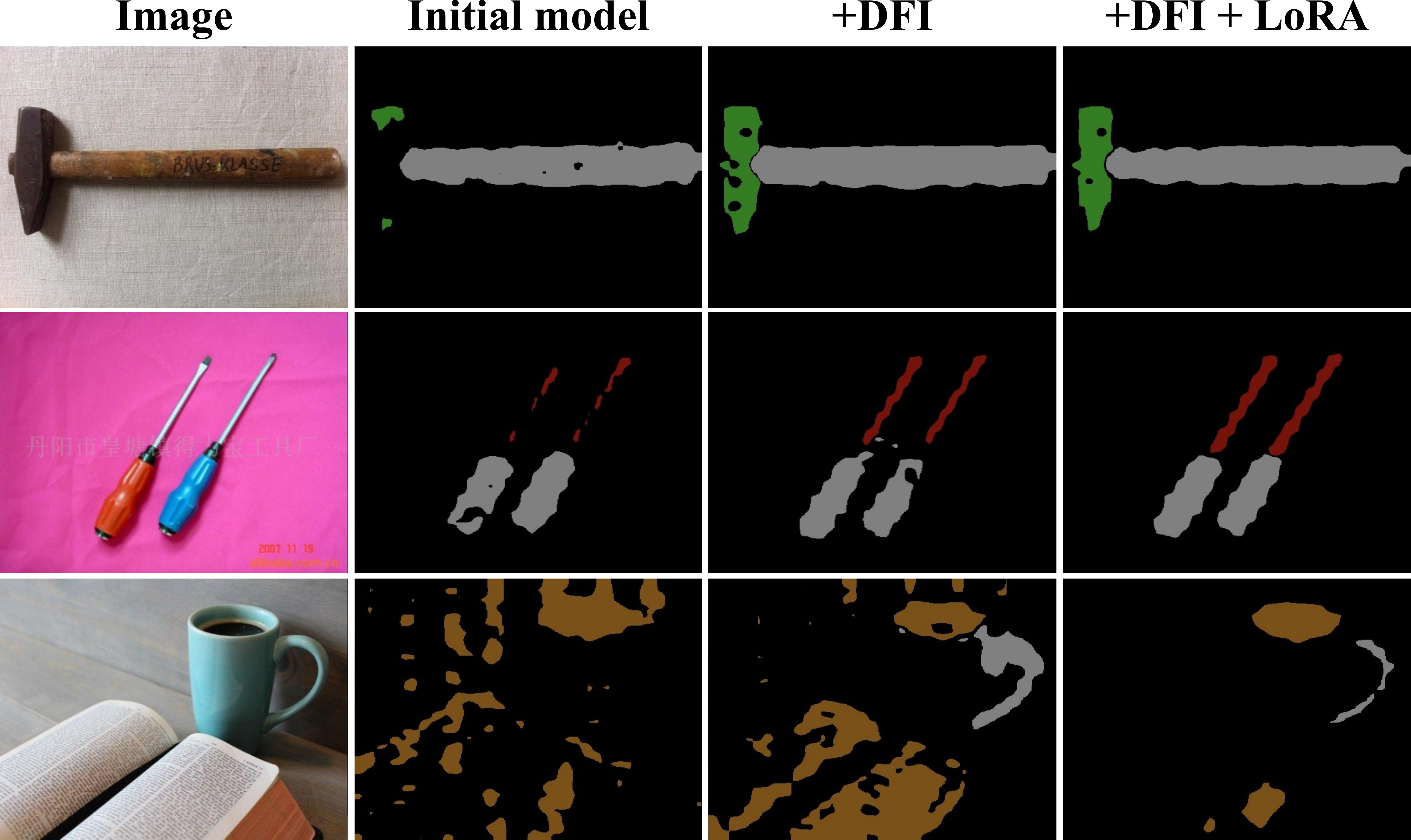}
\caption{Qualitative improvements with DFI and LoRA.}\label{fig:vis_ablations}
\vspace{-2mm}
\end{figure}
In particular, DFI can also be used solely in training and discarded at inference, improving results without extra computational cost.
The impact of this operation is analyzed in~\cref{tab:ablation-dfi-efficiency}, showing that the model efficiency can be improved when deactivating DFI at inference.
Additionally, integrating LoRA layers to fine-tune the foundation features is also helpful, leading to a 3.96\% improvement in mIoU with marginal additional parameters.

In~\cref{fig:vis_ablations}, we show the qualitative ablation results to visually examine the effects of DFI and LoRA.
The segmentation results indicate that DFI is particularly effective at locating tiny and slender parts, while LoRA further enhances performance with refined boundaries and more complete segmentation maps.
More ablation studies are provided in the supplementary.

\section{Conclusion}
In this paper, we present a streamlined affordance learning system that integrates automatic data collection from egocentric videos, effective model learning, and deployment on a real robot.
We perform extensive experiments from both vision and robot perspectives, demonstrating the effectiveness of the entire affordance learning system.

{
    \small
    \bibliographystyle{ieeenat_fullname}
    \bibliography{main}
}

\end{document}